\documentclass[conference]{IEEEtran}
\IEEEoverridecommandlockouts
\usepackage{cite}
\usepackage{amsmath,amssymb,amsfonts}
\usepackage{algorithmic}
\usepackage{graphicx}
\usepackage{textcomp}
\usepackage{xcolor}
\usepackage{float}
\usepackage{booktabs}
\def\BibTeX{{\rm B\kern-.05em{\sc i\kern-.025em b}\kern-.08em
    T\kern-.1667em\lower.7ex\hbox{E}\kern-.125emX}}
\begin{document}
\bstctlcite{IEEEexample:BSTcontrol}

\IEEEpubid{\makebox[\columnwidth]{979-8-3195-2794-3/26/\$31.00~\copyright 2026~IEEE\hfill}%
\hspace{\columnsep}\makebox[\columnwidth]{}}

\title{RadAI: Gaze-Intention Fusion Pipeline for Attention-Guided Educational Videos in Chest X-Ray Interpretation}

\author{
\begin{minipage}[t]{0.32\textwidth}\centering
Kavish Mehta\textsuperscript{1,3} \\
\textit{\textsuperscript{1}John Foster Dulles High School} \\
Houston, USA \\
kavishm.19@gmail.com
\end{minipage}\hfill
\begin{minipage}[t]{0.32\textwidth}\centering
Yuvan Siddharth Katamaneni\textsuperscript{2,3} \\
\textit{\textsuperscript{2}Tomball Memorial High School} \\
Houston, USA \\
yskatamaneni07@gmail.com
\end{minipage}\hfill
\begin{minipage}[t]{0.32\textwidth}\centering
Akash Awasthi\textsuperscript{3} \\
\textit{\textsuperscript{3}Houston Learning Algorithms Lab} \\
Univ. of Houston, Houston, USA \\
akashcseklu123@gmail.com
\end{minipage}
}

\maketitle

\begin{abstract}
Chest radiograph interpretation errors often arise from failures in visual search: although trainees may correctly identify an abnormality once pointed out, they frequently fail to fixate on the relevant region, while expert search strategies (where radiologists look, in what sequence, and for how long) remain largely implicit. Eye movement modeling examples (EMME), which replay a recorded expert's gaze over the image, have been shown to improve novice detection of findings on chest radiographs, but producing one requires an eye-tracking session on that specific case, so teaching sets are limited to recorded sessions by experts. In this work, RadAI is presented, the first end-to-end framework for transforming predicted radiologist gaze into narrated instructional videos, enabling scalable gaze-guided training without collecting new eye-tracking data. Given only a chest X-ray and its report, RadAI predicts an expert fixation sequence, associates each interval of that sequence with the diagnostic finding it pursues, extracts the corresponding image region, and renders the result as an EMME-style video in which each highlighted finding is narrated with synthesized speech. RadAI is evaluated on 97 cases from the EGD-CXR dataset~\cite{karargyris2021egdcxr}, assessing whether the highlighted regions correspond to annotated abnormality locations. We position this work as a framework for scalable generation of educational content for residents and medical students in the field of radiology. 
\end{abstract}

\begin{IEEEkeywords}
Biomedical imaging, computer aided instruction, gaze tracking, radiography, training, videos
\end{IEEEkeywords}

\section{Introduction}

\begin{figure*}[!t]
\centering
\includegraphics[width=\textwidth]{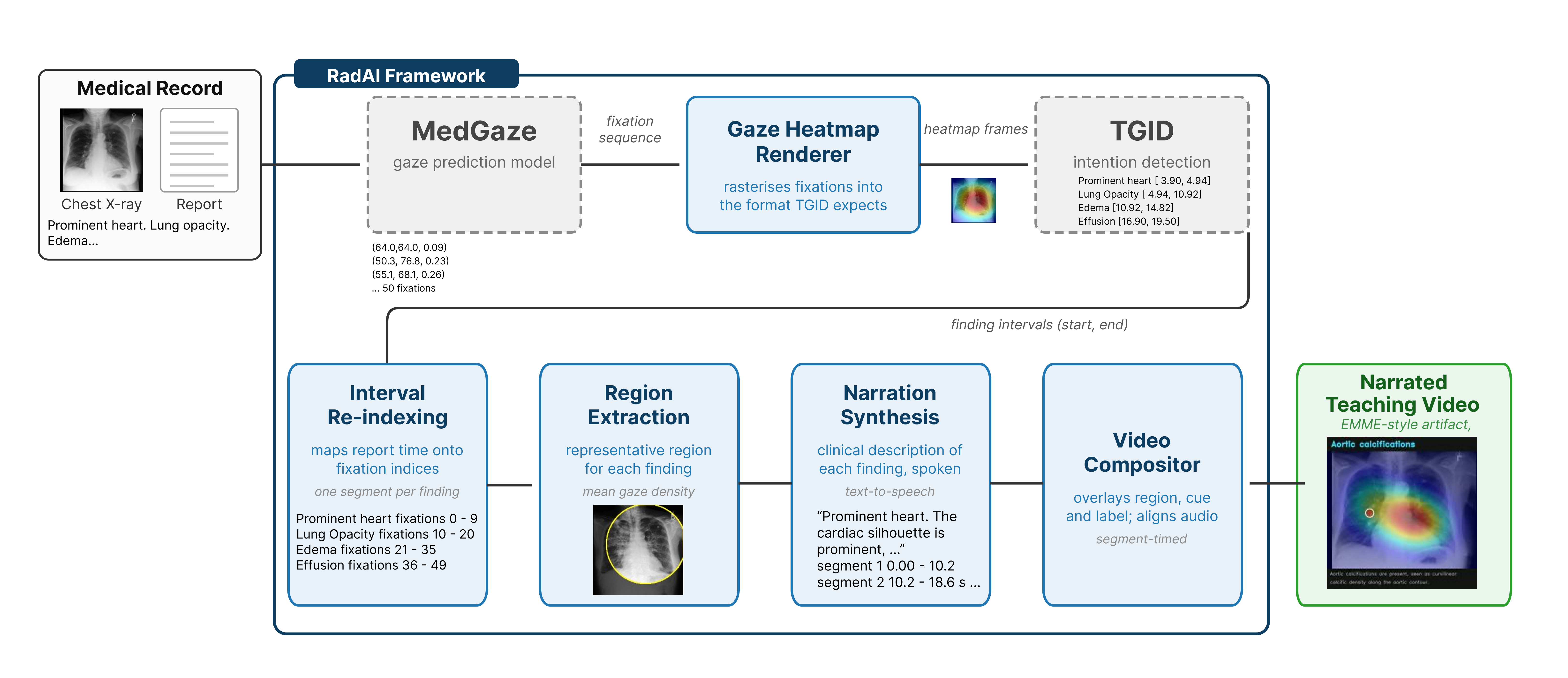}
\caption{The RadAI framework. A gaze prediction stage produces a fixation sequence from a chest radiograph and its report; an intention grounding stage associates each interval of that sequence with the diagnostic finding it pursues; a region extraction stage localizes each finding; and a rendering and narration stage assembles the labeled segments into a narrated instructional video.}
\label{fig:pipeline}
\end{figure*}

Chest radiography is among the most frequently performed imaging examinations, and interpretation errors remain common in this field. Studies of observer error in pulmonary nodule detection distinguish three main failures: the reader never fixates the lesion (search error), fixates it without recognizing it (recognition error), or recognizes it and dismisses it (decision error)~\cite{kundel1978visual}. Search errors account for roughly a third of misses, and they are the subset that attention guidance addresses. Expert search behavior differs substantially from that of novices~\cite{drew2013invisible, alexander2020radiologists}, but it is largely tacit: a radiologist can justify a diagnosis without being able to reconstruct the fixation sequence that produced it. Much of the effort to reduce radiological error with AI has targeted detection directly, building systems that find abnormalities on the reader's behalf. Search errors call for a different kind of intervention, one that trains where a reader looks rather than substituting the radiologist itself.

Teaching methods that combine visual and spoken instruction are well established in radiology education~\cite{mayer2021multimedia}. Eye movement modeling examples (EMME) replay a recorded expert's gaze superimposed on the stimulus, typically as a moving dot or heatmap, while the expert narrates their reasoning; learners who study EMME allocate their attention more like the model and interpret novel cases more accurately than learners who see the same material without the gaze overlay~\cite{jarodzka2013learning}. The effect has been demonstrated on chest radiographs specifically: novice radiographers shown another observer's recorded search located pulmonary nodules more reliably than under free search, with JAFROC detection scores rising from 0.56 to 0.60 after a single 20-second preview per image ~\cite{litchfield2010viewing}. EMME is therefore an answer to the problem posed in the previous paragraph: it teaches search directly, rather than substituting for it. The limitation however is not conceptual but logistical: producing one requires an eye-tracking session with an expert on that specific case, so a teaching set can only ever contain the recorded cases an expert was diagnosing.

Producing an EMME-style video without a recording requires more than any single existing model provides. The video must present a plausible expert gaze pattern; each portion of that pattern must be identified with the diagnostic finding it serves, since unexplained attention has no instructional value. The identified findings must be localized to image regions so the learner knows what to associate the narration with, and the whole video must be rendered in a form a learner can watch. Recent work supplies individual pieces of this chain. MedGaze predicts a radiologist's fixation coordinates and dwell times from a chest X-ray and its report, using a two-stage vision-language architecture that aligns visual features with report text before decoding the fixation sequence, and its predicted scanpaths have been judged by radiologists to resemble expert search~\cite{awasthi2025medgaze}. A temporally grounded intention detection (TGID) module, given a recorded eye-gaze heatmap video and its report, identifies which diagnostic finding the radiologist was pursuing during each interval, and a Region Extraction (RE) module isolates the corresponding image region~\cite{awasthi2024bridging}. No individual component, however, spans the full chain: gaze prediction alone yields attention without explanation, and intention detection alone assumes an existing visual input or recording.

In this paper, RadAI is introduced, the first end-to-end framework that transforms predicted radiologist gaze into narrated instructional videos. RadAI's design follows directly from the requirements above: a gaze prediction stage supplies the expert search pattern, an intention grounding stage attaches diagnostic meaning to each interval of it, a region extraction stage localizes each finding, and a rendering and narration stage assembles the result into an EMME-style video in which each finding is spoken aloud while its region is highlighted. The capability that emerges from this sequence exists in none of its parts: instructional gaze content can be generated for any case with a report on file, rather than only for the small set of cases on which an expert was recorded. This removes the central bottleneck of gaze-based radiology education — the per-case eye-tracking session — and makes expert-inspired instructional material producible at the scale of existing clinical archives. RadAI is evaluated on 97 cases from the EGD-CXR dataset, and we position this work as a framework for scalable educational content generation; establishing its effect on learner outcomes is a necessary next step that is discussed in Section~IV~.

\section{Methodology}

\subsection{RadAI Overview}

RadAI is organized as four sequential stages, shown in Fig.~\ref{fig:pipeline}. A gaze prediction stage takes a chest X-ray and its radiology report and produces a predicted fixation sequence. An intention grounding stage associates each interval of that sequence with the diagnostic finding pursued during it. A region extraction stage computes a representative image region for each finding. A rendering and narration stage assembles the findings, regions, and fixation sequence into a narrated instructional video. The first three stages build on existing models: MedGaze~\cite{awasthi2025medgaze} for gaze prediction, and the TGID and RE modules~\cite{awasthi2024bridging} for intention grounding and region extraction, while the whole end-to-end framework with rendering and narration is contributed by this work.

\subsection{Dataset}
All experiments use the EGD-CXR dataset~\cite{karargyris2021egdcxr}, which pairs chest radiographs with eye-tracking recordings and time-aligned report dictation collected while a radiologist read each case.

\subsection{Gaze Prediction Stage}

RadAI's input is a chest X-ray and its free-text radiology report. These are passed to MedGaze~\cite{awasthi2025medgaze}, which predicts an ordered sequence of fixations, each consisting of spatial coordinates $(x_i, y_i)$ and a dwell time $t_i$. MedGaze encodes the radiograph with a convolutional feature extractor followed by a transformer encoder. A query-based vision-language bridge then aligns the resulting visual features with the tokenized report and feeds a frozen language model. Fixation queries are decoded through Gaussian output heads over position and duration, yielding a fixed number of fixations per case. The publicly released model trained on EGD-CXR been used without modification or retraining. A property of its output shapes the design of the downstream stages: the predicted dwell times are relative quantities rather than calibrated absolute durations, which motivates the temporal normalization described in Section II-D.

\subsection{Intention Grounding and Interval Re-indexing}

The TGID module~\cite{awasthi2024bridging} predicts, for a gaze recording and its report, the set of diagnostic findings pursued during the recording and the start and end times of the interval associated with each finding. Its output is a list of (finding, interval) pairs on the timeline of the gaze sequence.

To place these intervals onto the predicted fixation sequence, RadAI constructs a temporal index over the MedGaze output: predicted dwell times are accumulated into a cumulative timeline and rescaled so that the final fixation coincides with the end of the report's dictation window. Each finding interval is then mapped to the set of predicted fixations whose rescaled cumulative times fall within it. The result is a segmentation of the predicted fixation sequence in which each segment carries a diagnostic label.

\subsection{Region Extraction}

For each labeled segment, a representative region is computed following the RE procedure~\cite{awasthi2024bridging}: the fixations belonging to the segment are accumulated into a single intensity map, with each fixation contributing a two-dimensional Gaussian centered at its coordinates and weighted by its dwell time. The accumulated map is normalized to $[0,1]$, and its high-intensity area constitutes the region associated with the finding. RadAI renders at $512 \times 512$ resolution with an isotropic Gaussian of standard deviation $\sigma = \text{150}$ pixels.

\subsection{Video Rendering and Narration}

The rendering stage converts the labeled segments into an instructional video. Each segment is rendered by color-mapping the accumulated intensity map and alpha-blending it over the radiograph. A marker then traverses the segment's fixations in their predicted order, and a caption bar displays the finding together with a one-sentence description. The description is drawn from a fixed template library keyed by finding and each description is synthesized to speech with gTTS.

Each segment's duration is governed by a temporal budget: it is set to the greater of the segment's summed predicted dwell time and the duration of its synthesized narration, so that neither the gaze presentation nor the audio is truncated. When the narration is the longer of the two, the frame allocation of individual fixations is scaled proportionally, preserving the relative dwell structure predicted by MedGaze. Segments are concatenated in report order, preceded by an uncluttered view of the radiograph, and encoded with the narration track as a single video. 

\begin{figure*}[!t]
\centering
\includegraphics[width=\textwidth]{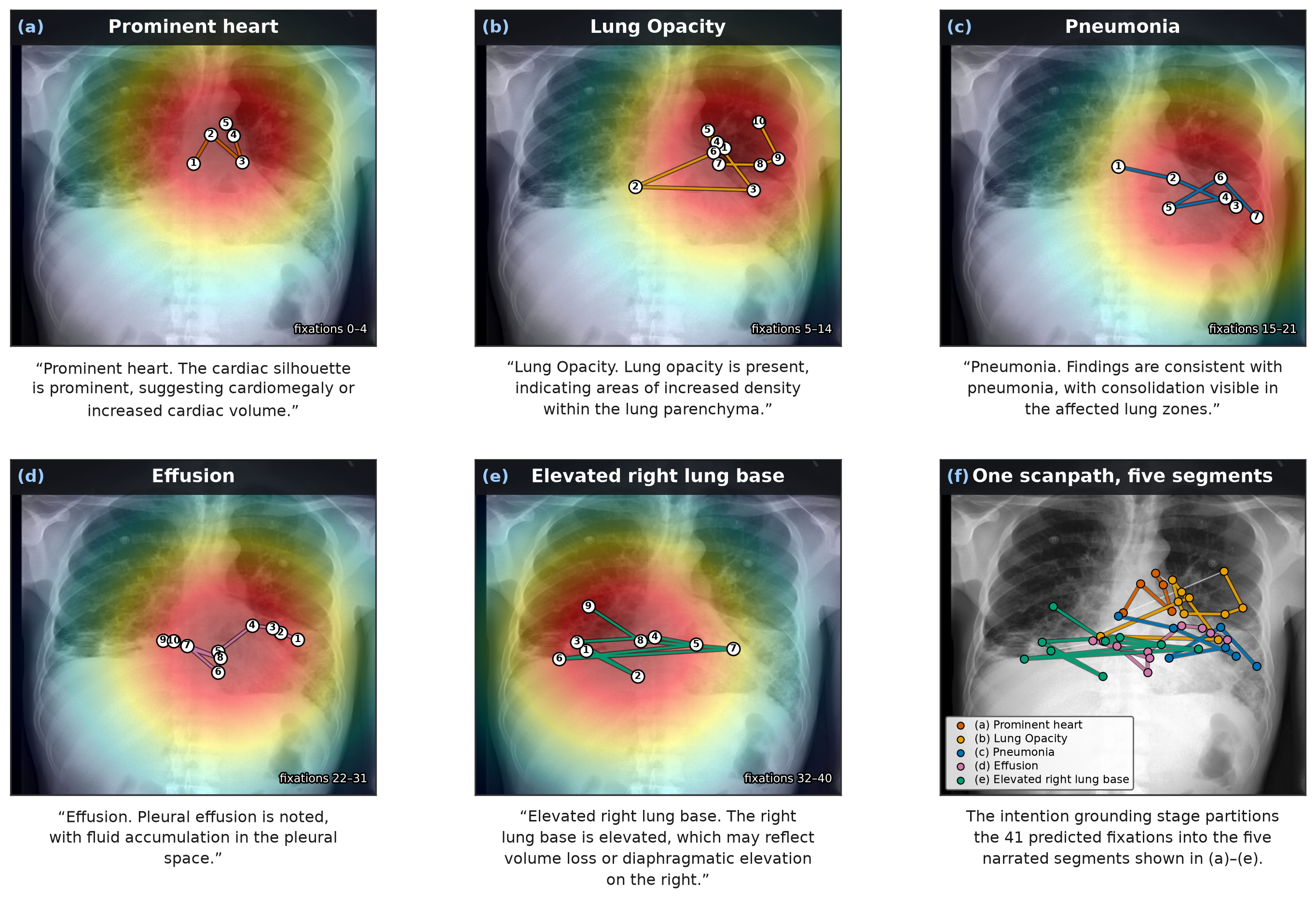}
\caption{Frames from a generated video for one case, one panel per narrated finding. Each finding panel shows that segment's accumulated attention heatmap over the radiograph, the predicted fixations numbered in the order the video visits them, and the sentence spoken while that segment is displayed; panels (a)--(e) appear in the video in the order shown. Panel (f) shows how the intention grounding stage partitions a single 41-fixation predicted scanpath into the five narrated segments.}
\label{fig:example}
\end{figure*}

\section{Experiments and Results}

\subsection{Setup and Implementation}

Instructional videos were generated using RadAI for 97 cases from the EGD-CXR test split, and the same cases are used for all evaluation. The gaze prediction stage uses a MedGaze checkpoint trained on EGD-CXR, applied without retraining or fine-tuning, and intention intervals are taken from TGID predictions on the corresponding recorded gaze videos. RadAI is implemented in Python using OpenCV for frame rendering and gTTS for narration synthesis; videos are encoded at 30 frames per second at $512 \times 512$ resolution. Fig.~\ref{fig:example} shows the frames of a generated video for one case. Across the 97 cases, RadAI produced 347 narrated segments, between 2 and 5 per video.

\subsection{Anatomical Correctness of the Demonstrated Search}

A generated video is only useful if the gaze it displays reaches the anatomy of the finding being narrated. For each narrated segment, the relevant anatomy is identified using a ChestX-Det segmentation model~\cite{lian2021structure, cohen2022torchxrayvision} --- the cardiac silhouette for cardiomegaly, the lung fields for opacity and pneumonia, the basal zones for effusion and atelectasis --- and record the segment as \emph{successful} when the majority of its dwell-weighted gaze falls inside that structure. We compare against two reference points: assigning each finding to a randomly chosen interval of the same scanpath, which is what the measure returns when the grounding carries no information, and substituting the true report timing for the predicted timing, which is the best any grounding could achieve given these annotations.

\begin{table}[H]
\caption{Success rate by finding}
\label{tab:findings}
\centering
\footnotesize
\begin{tabular}{lccc}
\toprule
Finding & $n$ & AOI area & Success \\
\midrule
Edema & 32 & 0.317 & 100\% \\
Pneumonia & 12 & 0.307 & 92\% \\
Lung Opacity & 33 & 0.331 & 88\% \\
Pneumothorax & 13 & 0.346 & 85\% \\
Atelectasis & 21 & 0.152 & 67\% \\
Cardiomegaly & 49 & 0.113 & 43\% \\
Effusion & 70 & 0.297 & 29\% \\
\midrule
Overall & 230 & 0.255 & 60\% \\
\bottomrule
\end{tabular}

\vspace{3pt}
\footnotesize\raggedright
AOI area is the fraction of the image occupied by that finding's anatomical target; smaller targets are correspondingly harder to land on.
\end{table}

Across the  narrated segments whose finding carries a ground-truth annotation, the demonstrated search is successful in \textbf{60\%} of segments. Random assignment succeeds in 46\%, confirming that the grounding contributes real information, and the true report timing succeeds in 57\% --- so the predicted grounding already performs at the level these annotations allow. Table~\ref{tab:findings} breaks this down by finding; the pattern is discussed in Section~IV.

\subsection{Conformance to Instructional Design Principles}

Because the output is a teaching artifact rather than a prediction, the video is also assessed against the multimedia design principles most associated with attention guidance~\cite{mayer2021multimedia}. The videos are segmented rather than continuous, presenting one finding at a time in units averaging 8.2 seconds within a mean total duration of 31.9 seconds, well inside the length at which viewer engagement is known to decline~\cite{guo2014video}. Narration sits at a Flesch-Kincaid grade level~\cite{kincaid1975derivation} of 13.5, appropriate for clinical trainees, and less than 1\% of runtime is silent.

\section{Discussion}

RadAI's central benefit is that it removes the per-case eye-tracking session from the production of gaze-guided teaching material. Recorded EMME requires an expert to be recorded reading the specific case being taught, which confines any teaching set to the cases that happened to be captured. RadAI requires only a radiograph and a report, both already present in the medical record, and produces a well-formed instructional video. That figure is the practical measure of the RadAI's reach: instructional content becomes producible at the scale of existing clinical archives rather than at the scale of eye-tracking studies.

The content it produces is also clinically grounded. In 60\% of narrated segments the gaze shown to the learner reaches the anatomy of the finding being described, against 46\% when findings are assigned to randomly chosen intervals of the same scanpath. More informative is the comparison against the ceiling: substituting the true report timing for the predicted timing succeeds in 57\%, so the predicted grounding is already performing at the level these annotations permit, and no further gain is available from better temporal alignment alone. The videos conform to the multimedia design principles most associated with attention guidance~\cite{mayer2021multimedia}, being segmented, appropriately paced in length, and pitched at a reading level suited to clinical trainees.

These results characterize what the framework produces, not what it teaches. Recorded EMME improves novice performance only when the demonstrated search is itself successful~\cite{litchfield2010viewing}; the analogous requirement for generated content is that the search a video displays actually find the finding it narrates, which is what the present evaluation tests. Meeting that condition is a precondition for instructional value, not a demonstration of it. Establishing that RadAI-generated videos improve trainee search behavior would require a controlled study with radiology learners of the kind used to validate recorded EMME~\cite{jarodzka2013learning}, which is outside the scope of this work.

Several further limitations bear on how these results should be read. RadAI's gaze predictions derive from a model trained on a single radiologist's recordings, so the search behavior it reproduces reflects one expert's style rather than a consensus of expert practice; whether this matters pedagogically is an open question, since instructional value may not require the demonstrated search to be modal. The narration is generated from a fixed template library keyed by finding, so it conveys what was found and where, but not the reasoning that connects the two --- recorded EMME derives part of its effect from the expert's spoken justification, which the present system does not reproduce. Finally, the highlighted regions are also less distinct from one another than would be ideal: cues for co-occurring findings overlap substantially, which limits how sharply a learner can associate each narration with a separate location. 

\section{Conclusion}

This work presented RadAI, an end-to-end framework that transforms predicted radiologist gaze into narrated instructional videos for chest radiograph interpretation. Given only a radiograph and its report, RadAI predicts an expert fixation sequence, associates each interval of that sequence with the diagnostic finding it pursues, extracts the corresponding image region, and renders the result as an EMME-style video in which each finding is narrated while its region is highlighted. Evaluated on 97 cases from EGD-CXR, the demonstrated search reached the anatomy of the narrated finding in 60\% of segments, against 46\% for a random-interval baseline and 57\% for the ground-truth report timing that bounds what these annotations allow.

The significance of the framework lies in what it removes rather than what it adds. Eye movement modeling examples are an established method for teaching visual search, but each one has required an eye-tracking session with an expert on the specific case being taught, which confines any teaching set to the cases that happened to be recorded. By generating the gaze sequence rather than recording it, RadAI makes instructional content producible for any case with a radiograph and a report on file --- that is, at the scale of existing clinical archives rather than at the scale of eye-tracking studies.

Two main directions follow directly from this work. The most immediate is a controlled study with radiology trainees to establish whether the generated videos affect search behavior and detection performance, which the present evaluation does not address. The second is extension beyond chest radiography to volumetric modalities such as CT, where search is conducted through a stack of slices and the temporal structure of expert attention is correspondingly richer.

\bibliographystyle{IEEEtran}
\bibliography{citations}
\end{document}